\pdfoutput=1
\documentclass[11pt,table]{article}

\usepackage[final]{acl}

\usepackage{times}
\usepackage{latexsym}

\usepackage[T1]{fontenc}
\usepackage[utf8]{inputenc}

\usepackage{microtype}

\usepackage{inconsolata}

\usepackage{graphicx}

\usepackage[utf8]{inputenc} % allow utf-8 input
\usepackage[T1]{fontenc}    % use 8-bit T1 fonts
\usepackage{hyperref}       % hyperlinks
\usepackage{url}            % simple URL typesetting
\usepackage{booktabs}       % professional-quality tables
\usepackage{amsfonts}       % blackboard math symbols
\usepackage{nicefrac}       % compact symbols for 1/2, etc.
\usepackage{microtype}      % microtypography

\usepackage{url}         
\usepackage{hyperref}    
\usepackage{graphicx}
\usepackage{xspace}
\usepackage{amsmath}
\usepackage[font=small]{caption}
\usepackage{algorithm}
\usepackage{algpseudocode}
\usepackage{comment}
\usepackage{pifont}
\usepackage{makecell}

\usepackage{tikz}
\usepackage{pgfplots}
\pgfplotsset{compat=1.18}
\usetikzlibrary{plotmarks}
\usepackage{tikz}
\usepackage{pgfplots}
\usepackage{tikz}
\usepackage{pgfplots}
\usepackage[utf8]{inputenc} 
\usepackage[T1]{fontenc}    

\pgfplotsset{compat=1.17} 

\usepackage{tabularx}
\usepackage{array}
\usepackage{ragged2e} 

\usepackage[many]{tcolorbox} 
\tcbuselibrary{skins}

\usetikzlibrary{calc}
\usetikzlibrary{positioning}
\usetikzlibrary{fit}

\usepackage{geometry}         
\usepackage{enumitem}         

\tcbuselibrary{breakable}
\usepackage{listings}  
\usepackage{textcomp}
\usepackage{stmaryrd}

\title{CoEx – Co-evolving World-model and Exploration}

\author{Minsoo Kim \\
   Seoul National University \\
  \texttt{minsoo9574@snu.ac.kr} \\\And
  Seung-won Hwang\textsuperscript{*} \\
  Seoul National University \\
  \texttt{seungwonh@snu.ac.kr} \\}

\begin{document}
\maketitle

\newcommand\blfootnote[1]{
  \begingroup
  \renewcommand\thefootnote{}\footnote{\hspace{-0.55\footnotesep}#1}%
  \addtocounter{footnote}{-1}
  \endgroup
}

\blfootnote{\textsuperscript{*}Corresponding author.}

\newcommand{\sw}[1]{\textcolor{orange}{\textbf{[SW]} }\textcolor{orange}{\textit{#1}}}
\newcommand{\mk}[1]{\textcolor{blue}{\textbf{[MK]} }\textcolor{blue}{\textit{#1}}}

\newcommand{\tabitem}{~~\llap{\textbullet}~~}

\newcommand{\ours}{CoEx\xspace}
\newcommand{\oursfull}{\textbf{Co}-evolving World-model and \textbf{Ex}ploration}

\definecolor{darkgreen}{rgb}{0.0, 0.5, 0.0}

\newcommand{\cmark}{\ding{51}}%
\newcommand{\xmark}{\ding{55}}%

%%%%%%%%%%%%%%%%
\newcommand{\env}{\text{Env}\xspace}
\newcommand{\planner}{\pi\xspace}
\newcommand{\actor}{\alpha\xspace}
\newcommand{\bk}{b_k\xspace}
\newcommand{\textmemk}{l_k\xspace}
\newcommand{\symbmemk}{m_k\xspace}
\newcommand{\plannerhk}{H_k\xspace}
\newcommand{\subgoalk}{e_k\xspace}
\newcommand{\subepisodetrajk}{\varepsilon_k\xspace}
\begin{abstract}
Planning in modern LLM agents relies on the utilization of LLM as an internal world model, acquired during pretraining.
However, existing agent designs fail to effectively assimilate new observations into dynamic updates of the world model.
This reliance on the LLM's static internal world model 
is progressively prone to misalignment
with the underlying true state of the world,
leading to the generation of divergent and erroneous plans.
We introduce a hierarchical agent architecture, \ours, 
in which hierarchical state abstraction allows LLM planning to co-evolve with a dynamically updated model of the world.
\ours plans and interacts with the world by using LLM reasoning to orchestrate dynamic plans consisting of subgoals,
and its learning mechanism continuously incorporates these subgoal experiences into a persistent world model in the form of a neurosymbolic \textit{belief state}, 
comprising
textual inferences and code-based symbolic memory.
We evaluate our agent across a diverse set of agent scenarios involving rich environments and complex tasks including ALFWorld, PDDL, and Jericho.
Our experiments show that \ours 
outperforms existing agent paradigms in planning and exploration.\footnote{Code will be publicly released after blind review.}

\end{abstract}

\section{Introduction}
\label{sec:intro}
While recent advances in large language model
(LLM) capabilities have enabled significant performance gains in LLM-based agents,
such agents fall short when  tasked with achieving goals in novel environments   with limited prior information. Meanwhile, humans particularly excel in such scenarios, actively \textbf{exploring} environments and \textbf{adapting} plans
based on new observations~\cite{Lake_Ullman_Tenenbaum_Gershman_2017,NEURIPS2018_2de5d166}.
With exploration,
the understanding of the environment, or \textit{world model}~\cite{NEURIPS2018_2de5d166,hafner2021mastering}, is continuously adapted, and planning is grounded on such understanding.

We attribute the limitations of existing LLM agents like
 ReAct~\cite{yao2023react} and Reflexion~\cite{shinn2023reflexion} in novel environments, to their \textbf{monolithic} design, which entangles planning, reasoning, and action generation within a single LLM instance. Instantiated through in-context learning (ICL) with action-level few-shot exemplars, this design faces two fundamental limitations:

\begin{itemize}
\item Exploitation Bias: ICL, driven by demonstrations of successful trajectories, biases the agent toward repeating known successful actions (i.e., exploitation) over exploration. With this bias, action-level planning results in a myopic horizon, limiting generalization to longer horizon tasks.

\item Limited Adaptation: The monolithic architecture complicates  integration of new exploratory insights into a persistent world model.
\end{itemize}

In contrast, we argue that agents need both the ability to plan \textbf{exploratory goals}, and the ability to directly leverage the outcome of exploration to update an \textbf{adaptive world model}.

We introduce our proposed framework, \oursfull~(\ours), which formulates LLM agent planning as a state machine over a subgoal-level belief state. 
Our first distinction is a Planner which conditions on a \textbf{subgoal-level} belief state, rather than low-level actions.
This enables \ours to
leverage subgoals to plan deliberate
exploration goals when faced with uncertainty.
An additional flexibility of our approach is dynamic planning based on exploration, 
where subgoals can be generated anew when the agent deems the current plan suboptimal.

Our second distinction is
 a robust mechanism for \textbf{world model co-evolution} through a process of verification and synthesis. Following each subgoal attempt, \ours distills task-relevant insights from raw experience, using LLM-based filtering to generate targeted updates to an adaptive belief state.

 Specifically, we propose a
 neurosymbolic belief state design, combining:
\begin{itemize}

\item Object-oriented symbolic memory for efficient low-level state tracking and,

\item LLM-based verification and synthesis for integrating new discoveries into the adaptive world model.
\end{itemize}

Through this unified subgoal-level planning and exploration strategy, \ours achieves co-evolving agent planning and adaptive world modeling, addressing the core limitations of monolithic LLM designs.
To demonstrate the effictiveness \ours, we evaluate our proposed method on a diverse set of challenging agent planning scenarios including ALFWorld~\cite{shridhar2021alfworld}, Jericho~\cite{Hausknecht_Ammanabrolu_Côté_Yuan_2020}, and PDDL~\cite{mcdermott_pddl_1998} domains.

\section{Related Work}
In Fig.\ref{fig:replanning_adaptation_landscape}, we illustrate the landscape of existing LLM agent paradigms along the dimensions of planning granularity and world model adaptation.
As overviewed in Section~\ref{sec:intro}, existing LLM agents 
with static world model and myoptic action-level planning 
fall in the \textbf{lower-left} (Fig.\ref{fig:replanning_adaptation_landscape}),
ReAct~\cite{yao2023react} and Reflexion~\cite{shinn2023reflexion}, relying on a monolithic agent design which conflates planning, reasoning, and action generation into a singular LLM agent.
ExpeL~\cite{Zhao_Huang_Xu_Lin_Liu_Huang_2024} implements an offline form of world model updates, but preclude real-time adaptation.
WALL-E~\cite{zhou2024walleworldalignmentrule} proposes offline rule learning for world model alignment, extracting symbolic rules from collected trajectories, but similarly does not support online adaptation.

Toward the \textbf{desired upper-right}, where ours is placed, existing efforts can be categorized
by two directions.
First, \textbf{world model update (upper left)}:
The upper-left shows approaches leveraging world model for planning,
repurposing LLMs as world models directly~\cite{hao-etal-2023-reasoning}, or using LLMs to generate code-based world models~\cite{tang2024worldcoder,dainese2024generating}.
In particular, the latter can represent adaptive knowledge as code~\cite{tang2024worldcoder,dainese2024generating} in an online manner.
However, by
LLM strictly taking the role of code generator,
LLM reasoning abilities cannot be leveraged for planning, instead delegating planning to external algorithms such as MCTS.  
In addition, they focus entirely on modeling action-level dynamics of the world, and in practice are limited to demonstrating basic competence in complex planning tasks such as AlfWorld~\cite{tang2024worldcoder}.

Second, \textbf{subgoal-level planning (lower right)}:
AdaPlanner~\cite{sun2023adaplanner} introduces dynamic plan improvement based on episodic and execution feedback, 
but rely on handcrafted examples to demonstrate plan adaptation in detail, and do not support learning with an explicit world model.
HiAgent~\cite{hu-etal-2025-hiagent}
employs subgoal-level planning, 
but similarly relies on implicit world models within monolithic agents,
using subgoals as memory chunks and observation summarization to reduce context redundancy.
Table~\ref{tab:agent_comparison} compares in  \ours with existing LLM agent frameworks in further detail.
\definecolor{halocolor}{HTML}{4285F4}       
\definecolor{reactcolor}{HTML}{EA4335}      
\definecolor{worldcodercolor}{HTML}{34A853} 
\definecolor{gifmctscolor}{HTML}{9C27B0}     
\definecolor{adaplannercolor}{HTML}{008080}  
\definecolor{hiagentcolor}{HTML}{9C27B0}  

\definecolor{awmcolor}{HTML}{FFA500}        
\definecolor{rapcolor}{HTML}{FFA500}         
\definecolor{reflexioncolor}{HTML}{FFA500}         
\definecolor{expelcolor}{HTML}{FFA500}         

\definecolor{wallecolor}{HTML}{34A853}         

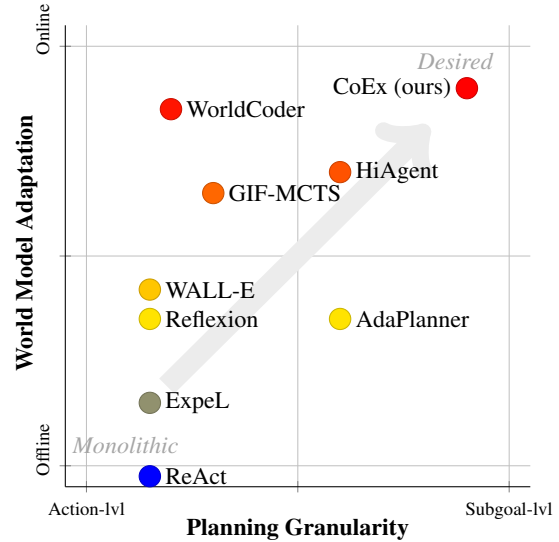
\begin{figure}[tbp]
\centering
\begin{tikzpicture}
\begin{axis}[
    width=1\linewidth,
    height=1\linewidth,
    xlabel={\textbf{Planning Granularity}},
    ylabel={\textbf{World Model Adaptation}},
    xlabel style={anchor=center,font=\small},
    ylabel style={rotate=0, anchor=center, xshift=1ex, font=\small},
    xticklabel style={font=\scriptsize},
    yticklabel style={font=\scriptsize,rotate=90, anchor=south east,xshift=4ex},
    xmin=-0.05, xmax=1.05,
    ymin=-0.05, ymax=1.05,
    xtick={0, 0.5, 1},
    ytick={0, 0.5, 1},
    xticklabels={Action-lvl, , Subgoal-lvl},
    yticklabels={Offline, , Online},
    tick align=outside,
    tick pos=left,
    grid=major,
    major grid style={line width=0.2pt, draw=gray!50},
    axis lines=left,
    axis line style={-},
    axis background/.style={fill=white},
]

\draw[->, very thick, gray!15, line width=8pt] (axis cs:0.19, 0.19) -- (axis cs:0.81, 0.81);

\addplot[scatter, only marks, mark=*, mark size=4pt, color=halocolor] coordinates {(0.9, 0.9)};

\addplot[scatter, only marks, mark=*, mark size=4pt, color=reactcolor] coordinates {(0.15, -0.025)};

\addplot[scatter, only marks, mark=*, mark size=4pt, color=worldcodercolor] coordinates {(0.2, 0.85)};

\addplot[scatter, only marks, mark=*, mark size=4pt, color=gifmctscolor] coordinates {(0.3, 0.65)};

\addplot[scatter, only marks, mark=*, mark size=4pt, color=adaplannercolor] coordinates {(0.6, 0.35)};
\addplot[scatter, only marks, mark=*, mark size=4pt, color=hiagentcolor] coordinates {(0.6, 0.7)};

\addplot[scatter, only marks, mark=*, mark size=4pt, color=reflexioncolor] coordinates {(0.15, 0.35)};

\addplot[scatter, only marks, mark=*, mark size=4pt, color=expelcolor] coordinates {(0.15, 0.15)};

\addplot[scatter, only marks, mark=*, mark size=4pt, color=wallecolor] coordinates {(0.15, 0.42)};

\node[align=center, font=\small] at (axis cs:0.5, -0.14) {Decoupled Planning $\rightarrow$ Full Replanning};
\node[align=center, font=\small, rotate=90] at (axis cs:-0.12, 0.5) {Static Model $\rightarrow$ Highly Adaptive};

\node[align=center, font=\footnotesize, text=gray!70] at (axis cs:0.10, 0.05) {
    \textit{Monolithic}
};

\node[align=center, font=\footnotesize, text=gray!70] at (axis cs:0.88, 0.965) {
    \textit{Desired}
};

\node[left=2pt, font=\small] at (axis cs:0.9, 0.9) {\ours (ours)};
\node[right=2pt, font=\small] at (axis cs:0.15, -0.025) {ReAct};
\node[right=2pt, font=\small] at (axis cs:0.2, 0.85) {WorldCoder};
\node[right=2pt, font=\small] at (axis cs:0.3, 0.65) {GIF-MCTS};
\node[right=2pt, font=\small] at (axis cs:0.6, 0.35) {AdaPlanner};
\node[right=2pt, font=\small] at (axis cs:0.6, 0.7) {HiAgent};
\node[right=2pt, font=\small] at (axis cs:0.15, 0.35) {Reflexion};
\node[right=2pt, font=\small] at (axis cs:0.15, 0.15) {ExpeL};
\node[right=2pt, font=\small] at (axis cs:0.15, 0.42) {WALL-E};

\end{axis}
\end{tikzpicture}
\caption{Landscape of LLM agent paradigms along the dimension of planning granularity (x-axis) ranging from action-level to subgoal-level, 
and frequency of world model adaptation (y-axis),
ranging from offline to online.
}
\label{fig:replanning_adaptation_landscape}
\end{figure}

\begin{table*}[th] 
\small
\centering 
\caption{\small Comparison of \ours with representative LLM agent architectures. 
}
\label{tab:agent_comparison}

\resizebox{\textwidth}{!}{ 
    \begin{tabular}{@{}l p{0.25\linewidth} p{0.25\linewidth} p{0.25\linewidth}@{}} 
    \toprule
    Method & World Model & Planning levels & Exploration Mechanism \\ 
    \midrule
    ReAct~\cite{yao2023react}     & Implicit    & Action-level  & \xmark  \\ 
    ExpeL~\cite{Zhao_Huang_Xu_Lin_Liu_Huang_2024}     & Offline Insights  & Action-level & \xmark     \\ 
    WALL-E~\cite{zhou2024walleworldalignmentrule}     & Offline Learned Rules  & Action-level & \xmark     \\ 
    Reflexion~\cite{shinn2023reflexion}     & Episodic Self-reflection    & Action-level & \xmark  \\ 
    AdaPlanner~\cite{sun2023adaplanner}  & Implicit  & Subgoal-level and Action-level (Monolithic LLM)   & \xmark  \\ 
    HiAgent~\cite{hu-etal-2025-hiagent}  & Implicit  & Subgoal-level and Action-level (Monolithic LLM)  & \xmark  \\    
    \midrule
    \ours (Ours)  & Adaptive Belief State updated after subgoal  & Subgoal-level (Planner) and Action-level (Actor) & Exploratory subgoals generated by Planner \\ 
    \bottomrule
    \end{tabular}
} 
\end{table*}

\section{Background}
In this section, we discuss the notions of exploitation bias and limited adaptation within the context of LLM agents.
We begin with a general reinforcement learning (RL) formulation of LLM agent's interaction with an environment, 
which can represent a wide range of interactive tasks with text-based observations~\cite{cote18textworld,Hausknecht_Ammanabrolu_Côté_Yuan_2020,li2024do}.

An LLM agent task can be defined as a
Partially Observable Markov Decision Process (POMDP), represented by the tuple
$(S, A, O, \operatorname{T}, \Omega, \operatorname{R}, \gamma)$,
where $S$ is the set of environment states $s$,
$A$ is the set of actions $a$,
and $O$ is the set of possible observations $o$.
$T$ is the transition probability between states, 
$\Omega$ is the observation probability, 
$R$ the reward function, and $\gamma$ the discount factor, respectively.

\subsection{Exploitation Bias: In-context Learning in LLM Agents}
In standard LLM agent paradigms such as ReAct~\cite{yao2023react} and Reflexion~\cite{shinn2023reflexion} 
the agent is defined by a singular action-level policy $\pi(a_t|o_t,h_t)$ where $a_t$ is the next action generated by the agent, given the latest observation $o_t$ and the trajectory history $h_t$ at timestep $t$.
This policy is instantiated 
through a process of in-context policy learning~\cite{shinn2023reflexion,monea2025llmsincontextbanditreinforcement}
from action-level exemplars demonstrating a reward-maximizing trajectory $\tau_j=(a_{j,0},o_{j,0},a_{j,1},o_{j,1}...a_{j,n},o_{j,n})$ where $\tau_j$ is the $j$th exemplar.

A shortcoming of in-context policy learning is ineffectiveness in exploration, biasing the agent policy toward exploitation
~\cite{monea2025llmsincontextbanditreinforcement,dai2024incontext}.
A primary reason behind this deficiency is an inability to learn from complex in-context signals~\cite{dai2024incontext}, 
which makes it difficult for the agent to acquire complex, long-horizon exploratory behaviors from action-level demonstrations.

In \ours, we overcome the exploitation bias of action-level ICL,
by instantiating exploratory behavior at the level of subgoals,
decoupling exploration from being dependent on the action-level demonstrations.
Instead of needing to plan exploratory action sequences, our approach allows exploration to be easily orchestrated at an abstracct level, as a standard subgoal of the Planner.

\subsection{Proposed World Model: Adaptive Belief State} 
In RL, the notion of \textit{belief states} is introduced as a solution to partial observability.
A key characteristic of POMDPs is the partial observability of the true $s_t$. 
That is, the observable state $o_t$ is an imperfect description of $s_t$,
and the belief state $b$ acts as a sufficient estimate of the underlying $s_t$~\cite{DBLP:journals/corr/HausknechtS15,avalos2024the},
that can be modeled by the agent~\cite{NEURIPS2020_1fc30b9d}.
Within the context of LLM agents in POMDPs, 
the notion of belief state is functionally equivalent to the definition of an LLM's world model,
under the definition of world model as a \textit{mechanism by which the LLM agent can estimate} some \textit{task-sufficient world state} $s^*_t$, 
where $s^*_t$ encapsulates the minimal set of variables or properties of the world state that are necessary for optimal LLM planning towards the task goal~\cite{li2024do}; That is, $s^*_t$ and the belief state $b_t$ are equivalent.

Given this, we can pinpoint a shortcoming of the monolithic LLM agent paradigm.
As monolithic LLM agent's world model is 
implicit in the LLM agent's parametric encoding of the trajectory history $h_t$,
$s^*_t$ cannot be reliably adapted in response to experiences, including exploratory ones.
Given an implicit world model, controlled updates to it are difficult,
and the implication of this is that, in POMDPs, the monolithic LLM agent cannot reliably leverage world model adaptation to resolve the uncertainty arising from partial observability.

In \ours, we address this problem by defining an explicit belief state as an adaptable world model,
whose formulation allows for controllable, targeted updates to the LLM's world model.
In order to facilitate efficient adaptation, we leverage a neurosymbolic design which integrates experiences at the subgoal level,
which also neatly dovetails with subgoal-level organization of planning.

\section{Method}
\subsection{Architecture}
\ours addresses these challenges through the co-evolution of world model and exploration, 
leveraging a hierarchical architecture consisting of the following components, each of which will be described in detail in subsequent sections. We highlight the distinction of ours in bold:

\paragraph{Planner (\ref{sec:method_subgoal_planner})} An LLM agent, implemented at the
\textbf{subgoal-level}, which reasons and plans the task through the generation of abstract subgoals.
\paragraph{Actor (\ref{sec:method_actor})} An LLM agent, whose 
policy designed to efficiently carry out subgoals in the environment, through a reasoning-and-acting loop.
\paragraph{Adaptive Belief State (\ref{sec:method_adaptive_beliefstate})} An adaptable, explicit belief state representation comprised of a code-based, object-oriented symbolic memory, and a structured textual memory generated using LLM-based verification and synthesis module, as $\bk=(\symbmemk, \textmemk)$.

Our approach integrates these components into a unified loop of planning, exploration, and learning from experience.
The full operation of \ours is presented in Alg.\ref{alg:mainalgo}, which we describe in detail below.

\begin{table}[t]
    \centering
    \small 
    \renewcommand{\arraystretch}{1.4} 
    \begin{tabularx}{\linewidth}{| >{\RaggedRight\arraybackslash}X |}
    \hline

    \multicolumn{1}{|>{\Centering\arraybackslash}X|}{\textbf{Planner Step:} k=1} \\
    \hline
    \textbf{Goal}: You need to find your way into a secret passage where the entrance is in the living room of the house.
    \par\vspace{0.5em}
    \textbf{Initial Observation}: You are standing in an open field west of a white house, with a boarded front door. There is a small mailbox here.     
    \\ \hline
    
    \textbf{Planner Reasoning}: To achieve the goal of finding the secret passage in the living room, I need to first enter the house. Since the front door is boarded, I should check the mailbox for any clues or items that might help me. 
    \\ \hline
    \textbf{Plan Subgoals:} \par
    1. Examine the mailbox. \par
    2. Look for an alternative way to enter the house (e.g., a window or back door). \par
    3. Enter the house. \par
    4. Navigate to the living room. \par
    5. Find the secret passage.
    \\ \hline 
    \textbf{Execute Subgoal:} [Examine the mailbox.] 
    \\ \hline 

    \multicolumn{1}{|>{\Centering\itshape\arraybackslash}X|}{
        (\dots Planner history $\plannerhk$ and belief state $\bk$)
    } \\ 

    \hline 
    \multicolumn{1}{|>{\Centering\arraybackslash}X|}{\textbf{Planner Step:} k=5} \\
    \hline
    \textbf{Planner Reasoning}: I have successfully navigated to the living room. Now, I need to find the secret passage. In the living room, I have noticed a trap door, which could potentially lead to the secret passage.
    \\ \hline
    \textbf{Plan Subgoals:} \par
    1. Investigate the trap door in the living room. \par
    2. Find the secret passage.
    \\ \hline
    \textbf{Execute Subgoal:} [Investigate the trap door in the living room.] %\par
    \\ \hline
    \end{tabularx}
  \caption{Illustration of \ours's subgoal-level planning in the classical Jericho game, \textit{Zork1}.}
  \label{tab:jericho_plan_example_revised_vertical}
\end{table}
\subsection{Subgoal Planner}
\label{sec:method_subgoal_planner}
\label{sec:method_wm_conditioned_planning}
\ours's Planner $\planner$ functions as the primary driver of task as well as world exploration progress. 
Its implementation enables the application of LLM reasoning towards robust subgoal-level planning. Our planner is an LLM prompted as a state machine,
where the states are defined by the neurosymbolic belief state $b_k$, 
and its action space consists of \textit{subgoal execution commands}, which allows the planner to request any subgoal it can describe via natural language.
The planner, given its context history $\plannerhk$, which consists of past belief states, generated subgoals, and associated reasoning, generates a new subgoal by reasoning about this history and the latest belief state $b_k$ (Alg.\ref{alg:mainalgo},\textit{Ln.}~\ref{lst:line:subgoal_gen}).

In Table~\ref{tab:jericho_plan_example_revised_vertical}, we illustrate the subgoal planning procedure using an example from the \texttt{Zork1} of the Jericho~\cite{Hausknecht_Ammanabrolu_Côté_Yuan_2020} text-based game suite.
Given the goal \textit{"You need to find your way into a secret passage where the entrance is in the living room of the house."}, and initial observation, $\planner$ reasons about the task and generates an initial plan consisting of subgoal steps.
As it proceeds through the task, it adapts its plan to the evolving $b_k$, as demonstrated in the later planner step $k=5$.
The abstraction of subgoal-level planning as \textit{reasoning over a belief state} $b_k$,
distinct from action-level planning,
allows the planner
to exhibit a high degree of flexibility in planning, 
overcoming exploitation bias in action-level ICL.

We demonstrate two key properties of our planner in this example, 
1) Dynamic Replanning: Our planner can not only generate entire plans for a task and drive the task by progressing through each subgoal, but can dynamically adjust the plan in response to the evolving belief state at each timestep $k$,
as illustrated in its generation of an updated plan \textit{1. Investigate the trap door in the living room., 2. Find the secret passage.}\footnote{We instruct the planner to generate plans using a \texttt{FULL PLAN} tag, followed by enumerated subgoals of the plan. Detailed prompts are provided in Appendix~\ref{appendix:planner_prompts}.}
2) Exploration and adaptation: As the task goal never actually specifies that the secret passage is behind the trap door, solving the task requires exploring the world and adapting to new discoveries.
In this example, \ours adapts to the discovery of a trap door,
and proceeds to explore the world further using the exploratory subgoal \textit{Investigate the trap door in the living room}.
As subgoals allow $\planner$ to easily organize exploration as discrete objectives,
our approach enables strategic and systemic exploration.
As a result, \ours's subgoal planning, leveraging powerful LLM reasoning on belief states,
facilitates both task progression and exploration within a unified framework.

\subsection{Actor}
\label{sec:method_actor}
The Actor $\actor$ is responsible for carrying out the subgoals
according to the specification generated by the planner. 
It generates the low-level actions to interact with the environment using a reason-and-act loop,
with the explicit objective of completing the assigned subgoal (\textit{Ln}.~\ref{lst:line:subgoal_episode}).

To design an effective subgoal-level Actor, we address two critical challenges:
1) The subgoal-level actor needs to handle potentially diverse subgoals requested by the planner.
To address this, we prompt the actor using a shared skill library~\cite{khot2023decomposed},
which can be dynamically leveraged by the Actor depending on the subgoal.
We generate a library of subgoal exemplars by decomposing existing task-level exemplars into smaller chunks, 
and annotating the smaller subtrajectories with an appropriate subgoal.
For example, in the PDDL task \texttt{gripper}, 
we decompose a task-level trajectory into skills which include \textit{pick up object}, \textit{move to room}, \textit{drop object}, etc.

2) To effectively collaborate with the Planner,
the actor needs to generate a self-contained subgoal-execution episode, which efficiently attempts the subgoal and does not continue indefinitely if the subgoal cannot be completed.
As a solution, we prompt the actor with instructions which enable it to self-judge the status of the subgoal-execution sub-episode,
terminating if the subgoal is completed, is deemed unachievable, or if the sub-episode exceeds a set number of steps.\footnote{We set the number of steps to 35.}
This design allows the actor to generate self-contained subgoal-execution trajectories, as well as flexibly pass control back to the planner.
The actor's algorithm is shown in (Alg.\ref{alg:hierarchical_hypoagent_direct_subgoal}, \textit{Ln}.~\ref{lst:line:subgoal_execution}), and example prompts are shown in Appendix~\ref{appendix:subgoal_prompts}.

\newcommand{\symbolicMemoryContent}{
    \textbf{[Agent]}\par
    Location: at Living room\par
    Inventory:\par
    - Obj: brown sack\par
    - Obj: clove garlic\par
    \medskip 
    
    \textbf{[Visited Locations]}\par
    - Loc: Attic\par
    - Loc: Behind house\par
    - Loc: Clearing\par
    - Loc: Forest\par
    - Loc: Kitchen\par
    \dots
    \medskip 
    
    \textbf{[Discovered Objects]}\par
    - Obj: clove garlic (at: inventory)\par
    - Obj: brown sack (at: inventory)\par
    - Obj: kitchen table (at: Kitchen)\par
    - Obj: ancient map (at: Living room)\par
    - Obj: brass lantern (at: Living room)\par
    \dots
    \par
    - Obj: trap door (at: Living room)
}

\newcommand{\structuredMemoryContent}{
    \textbf{[Current Plan]}\par
    Subgoals\par
    1. Navigate to the living room.\par
    2. Find the secret passage.\par
    Status: Successfully navigated to the living room, progressing towards finding the secret passage.\par
    \medskip 
    \textbf{[Subgoal Verification]}\par
    - Description: Navigate to the living room\par
    - Outcome: Completed\par
    - Justification: The last subgoal of navigating to the living room was completed successfully, which is a direct step towards the next subgoal of finding the secret passage.\par
    \medskip
    \textbf{[Learned Facts]}\par
    - Error: Agent encountered a game error when attempting to go north, indicating a possible dead end in that direction.\par
    \dots
}

\begin{table}[t!] 
    \centering
    \small 
    \renewcommand{\arraystretch}{1.3} 
    \begin{tabularx}{\linewidth}{| >{\RaggedRight\arraybackslash}X | >{\RaggedRight\arraybackslash}X |}
    \hline
    \multicolumn{2}{|c|}{\textbf{Belief State ($\bk$)}} \\
    \hline
    \multicolumn{1}{|c|}{\textbf{\begin{tabular}[c]{@{}c@{}}Object-oriented\\Symbolic Memory ($\symbmemk$)\end{tabular}}} &
    \multicolumn{1}{|c|}{\textbf{\begin{tabular}[c]{@{}c@{}}Structured\\Text Memory ($\textmemk$)\end{tabular}}} \\
    \hline
    \symbolicMemoryContent 
    &
    \structuredMemoryContent 
    \\ \hline
    \end{tabularx}
  \caption{Illustration of a belief state in \texttt{Zork1}, consisting of symbolic and structured text memory. }
  \label{tab:belief_state_components}
\end{table}
\subsection{Adaptive Belief State}
\label{sec:method_adaptive_beliefstate}
The key to our method's 
world model adaptation is the explicit belief state $b_k$, which evolves as subgoals are planned and executed.
The belief state is updated using the subgoal experience $\subepisodetrajk$, through two complementary representations, $\symbmemk$ and $\textmemk$.
Our dual-representation design targets fast, symbolic updates for low-level state tracking,
while leveraging more powerful LLM inference to augment the symbolic state with structured inferences.

\paragraph{Symbolic Memory ($\symbmemk$)}
We draw from the notion of object-oriented representation in RL~\cite{10.1145/1390156.1390187}, 
to focus the symbolic representation around objects and ego-centric agent
information,
generally applicable to wide array of agent scenarios.
$\symbmemk$ is implemented as a code-based, object-oriented representation storing concrete, factual information (e.g. agent location, object states, locations), derived from the raw trajectory $\subepisodetrajk$. 
An example of the symbolic memory is shown in the left column of Table~\ref{tab:belief_state_components}.

The symbolic memory is designed associated programmatic update mechanism, that systematically translates low-level actions $a_t$ and corresponding observations $o_t$ from interaction, into updates of the symbolic representation.
Implementing symbolic memory in code allows for fast, efficient state tracking, deliberately offloading the burden of lower-level state tracking processing from the planner $\planner$. 
We update the symbolic memory at each timestep $t$ during subgoal execution.\footnote{The implementation details and code of the symbolic memory can be found in the Appendix~\ref{sec:symbmem_details}.}

\paragraph{Structured Textual Memory ($\textmemk$)}
Complementing the fast, factual tracking in $\symbmemk$, the structured textual memory $\textmemk$ aims to capture the agent's higher-level understanding, 
manages uncertainty, 
tracks task progress, 
and synthesizes knowledge that may be less amenable to rigid symbolic representation, 
using natural language.

Distinctly from $\symbmemk$, the structured textual memory is updated through a 2-stage verification and synthesis process, evaluating the subgoal execution outcome $\symbmemk$ and $\subepisodetrajk$ before incorporating information into the belief state, using an LLM-based Verification and Synthesis module $v$.
An example of the resulting structured textual memory is shown in the right column of Table~\ref{tab:belief_state_components}.

\paragraph{Stage 1: Verification} The first stage is verification, where the module analyzes the trajectory $\epsilon_k$ and the updated symbolic state $m_{k+1}$ through a series of queries, 
to assess consistency, detect execution failures or unexpected events, and determine the success or progress of the subgoal.
To facilitate the incorporation unexpected discoveries,
we also include a question about new facts learned, or surprising outcomes.

\paragraph{Stage 2: Belief Synthesis} 
\label{sec:synthesis}
Based on the verification QA results and $\epsilon_k$ and $m_{k+1}$ as well as previous belief state, the next stage synthesizes the structured updates to the belief state:
We use an LLM to generate the \texttt{status line}, which reflects the last subgoal outcome,  
the \texttt{justification}, the rationale for the status line, 
and \texttt{learned facts}, which act as a storage for new facts and hypotheses formed.
The generated output is is the new $\textmemk$.

Combined together, the updated $\symbmemk$ and $\textmemk$ together constitute the new $\bk$.
We show the prompts of the Verification and Synthesis modules in Appendix~\ref{appendix:vas_prompts}.
\section{Experiments}
\begin{table*}[ht]
\centering
\tiny
\caption{Results on the six tasks of the ALFWorld benchmark, using success rate (\%) as evaluation metric. Best results are highlighted in bold.
}
\label{tab:alfworld_detailed}
\resizebox{0.9\linewidth}{!}{
\begin{tabular}{l | c c c c c c | c}
\toprule
Method & Pick & Clean & Heat & Cool & Examine & Picktwo & Total \\
\midrule
ExpeL & - & - & - & - & - & - & 64.20\% \\
ReAct & 66.67\% & 41.94\% & 91.03\% & 80.95\% & 55.56\% & 35.29\% & 61.94\% \\
Reflexion & 75.00\% & 90.32\% & 91.30\% & 90.48\% & 88.89\% & 94.12\% & 88.06\% \\
AdaPlanner & \textbf{100.00}\% & 96.77\% & 95.65\% & \textbf{100.00}\% & \textbf{100.00}\% & 47.06\% & 91.79\% \\
WALL-E & \textbf{100.00}\% & \textbf{97.00}\% & \textbf{100.00}\% & 86.00\% & 85.00\% & \textbf{100.00}\% & \textbf{95.00}\% \\
\midrule
\ours (Ours) & \textbf{100.00\%} & 83.87\% & 91.3\% & 90.48\% & \textbf{100.00\%} & 88.24\% & 93.28\% \\
\bottomrule
\end{tabular}
}
\end{table*}
\begin{table}[ht]
\centering
\tiny
\caption{Results on the PDDL task domains,
using success rate (sr\%) and progress rate (pr\%) as evaluation metric. Progress rate measures subgoal completion progress, separately from task success.
}
\label{tab:pddl_results}
\resizebox{1\linewidth}{!}{
\begin{tabular}{l | c c | c} 
\toprule
 & Gripper & Blocksworld & Average \\
Method & (sr/pr \%) & (sr/pr \%) & (sr/pr \%) \\
\midrule
ReAct & 65.0\%/89.5\% & 50.0\%/65.0\% & 60.0\%/81.3\% \\ 
\midrule
HiAgent & \textbf{75.0}\%/89.9\% & 50.0\%/68.3\% & 66.7\%/82.7\% \\ 
\midrule
\ours (Ours) & 70.0\%/\textbf{94.2\%} & \textbf{80.0\%/90.0\%} & \textbf{73.3\%/92.8\%} \\ 
\bottomrule
\end{tabular}
}
\end{table}
\definecolor{easycellcolor}{RGB}{220,255,220} 
\definecolor{hardcellcolor}{RGB}{255,220,220} 
\begin{table}[!ht]
\centering
\small
\caption{Results on the Jericho text adventure games, using success rate (sr\%) and progress rate (pr\%) as evaluation metrics.
Games are color-coded by difficulty: \colorbox{easycellcolor}{easy} and \colorbox{hardcellcolor}{hard}.
}
\label{tab:jericho_detailed}
\resizebox{0.9\linewidth}{!}{
\begin{tabular}{p{3.2cm}|c|c}
\toprule
& ReAct & \ours (Ours) \\
Game & pr\% & pr\% \\
\midrule
\cellcolor{hardcellcolor}905 & 0.0 & 85.7 \\
\cellcolor{hardcellcolor}acorncourt & 9.1 & 45.5 \\
\cellcolor{easycellcolor}afflicted & 0.0 & 100.0 \\
\cellcolor{hardcellcolor}balances & 42.9 & 42.9 \\
\cellcolor{easycellcolor}dragon & 100.0 & 100.0 \\
\cellcolor{easycellcolor}jewel & 66.7 & 0.0 \\
\cellcolor{easycellcolor}library & 25.0 & 25.0 \\
\cellcolor{easycellcolor}omniquest & 25.0 & 100.0 \\
\cellcolor{easycellcolor}reverb & 50.0 & 75.0 \\
\cellcolor{easycellcolor}snacktime & 100.0 & 100.0 \\
\cellcolor{hardcellcolor}zenon & 16.7 & 50.0 \\
\cellcolor{hardcellcolor}zork1 & 50.0 & 100.0 \\
\cellcolor{easycellcolor}zork2 & 50.0 & 50.0 \\
\cellcolor{easycellcolor}zork3 & 25.0 & 25.0 \\
\cellcolor{hardcellcolor}detective & 0.0 & 0.0 \\
\cellcolor{hardcellcolor}night & 0.0 & 0.0 \\
\cellcolor{hardcellcolor}pentari & 83.3 & 60.0 \\
\cellcolor{hardcellcolor}weapon & 66.7 & 50.0 \\
\cellcolor{easycellcolor}huntdark & 0.0 & 66.7 \\
\cellcolor{hardcellcolor}loose & 16.7 & 33.3 \\
\midrule
Avg. pr\% (Easy) & 35.2 & \textbf{61.9} \\
Avg. pr\% (Hard) & 37.4 & \textbf{50.7} \\
Avg. pr\% (All) & 36.4 & \textbf{55.5} \\
Overall sr\% & 10.0 & \textbf{25.0} \\
\bottomrule
\end{tabular}
}
\end{table}

\subsection{Experimental Setup}

\paragraph{Benchmarks} 
We evaluate \ours across 3 distinct agent benchmarks evaluating aspects of planning, exploration and world modeling.
\textbf{ALFWorld}~\cite{shridhar2021alfworld} is a text-based embodied task, requiring grounding to household environments and multi-step execution. We follow baselines and evaluate on the standard unseen test split.
\textbf{Jericho}~\cite{Hausknecht_Ammanabrolu_Côté_Yuan_2020} is a suite of text-based adventure games requiring exploration, world modeling, and common-sense reasoning to achieve high scores.
We adopt the setting from AgentBoard~\cite{agentboard}, which converts the games from open-ended exploration without a specified goal, to a task with a human-annotated goal (e.g. "Get out of the house. Then escape the city without getting caught via driving.").
We also evaluate on \textbf{PDDL}~\cite{mcdermott_pddl_1998}, 
a classical symbolic planning domain testing logical planning and state tracking,
using the Gripper and Blocksworld domains involving complex multi-step robotic planning.

\paragraph{Baselines} We compare \ours against representave LLM agent architectures for planning.
ReAct~\cite{yao2023react} is a canonical model-free baseline using a think-act loop. 
Reflexion~\cite{shinn2023reflexion} extends ReAct with self-reflection for error correction based on past trials, implementing a low-frequency form of world model adaptation.
AdaPlanner~\cite{sun2023adaplanner} is an agent that adapts its plan based on execution feedback, enabling dynamic replanning through detailed replanning exemplars.
HiAgent~\cite{hu-etal-2025-hiagent} is an agent which employs subgoal-based planning by leveraging subgoal observation chunk summarization, to focus on subgoal-relevant contexts.

ExpeL~\cite{Zhao_Huang_Xu_Lin_Liu_Huang_2024} implements a form of offline world model adaptation, 
learning static beliefs about the task through offline learning on success and failure trajectories.
WALL-E~\cite{zhou2024walleworldalignmentrule} also leverages offline rule learning from collected trajectories for world model alignment, but leverages code-based symbolic rules.

\paragraph{Implementation Details}
We set a maximimum number of total steps taken in each environment at 100, 100 and 150 for ALFWorld, PDDL, and Jericho, respectively. We set the maximum number of subgoal execution steps by the actor, at 35.
We utilize GPT-4o-mini for all experiments, for all components of \ours. 
For all prompting, we utilize a system prompt followed by an instance prompt. The details of prompts can be found in Appendix~\ref{appendix:prompts}.

\subsection{Results}
\paragraph{ALFworld} We report the results of experiments on ALFWorld in Table~\ref{tab:alfworld_detailed}.
\ours demonstrates strong performance across all task types, achieving the highest total success rate of $93.28\%$,
over state-of-the-art baselines.
ALFWorld tasks require a mixture of planning and exploration,
as the tasks require locating objects whose locations are at first unknown,
then carrying out further tasks involving them.
Through its co-evolution of planning and world model adaptation,
\ours achieves superior performance compared to less dynamic methods for world model adaptation, such as ExpeL or Reflexion.
It also achieves competitive performance with WALL-E, an offline world-model learning approach that requires a dedicated offline learning phase to generate rules across multiple episodes, demonstrating the efficiency of \ours's online learning approach.
Compared to AdaPlanner, which implements a strong dynamic planning approach,
our method demonstrates better overall performance as well as significantly higher performance on the more challenging "picktwo" task type, involving locating two distinct objects and placing them in the correct target locations.

\paragraph{PDDL} We compare \ours against ReAct on PDDL domains, and report the results in Table~\ref{tab:pddl_results}.
As PDDL tasks generally involve more numbers of subgoals than ALFWorld, we adopt the progress rate metric from \citet{agentboard} which allows tracking subgoal progress rate independently from task success rate.
\ours outperforms ReAct by 13.3\%/11.5\% in both success rate and progress rate, 
demonstrate the strong capability of 
our subgoal-based approach to LLM agent planning in complex planning domains.

\paragraph{Jericho} We report the results on the Jericho text-based games in Table~\ref{tab:jericho_detailed}.
Jericho games require strong exploration capabilities, as significant amounts of information about the environment are only attainable through learning from exploration.
The results demonstrate that \ours excels over ReAct in exploration, showing a 15\%/19.1\% gain in success rate and progress rate, respectively.
Following \citet{agentboard}, we further divide the games into easy and hard difficulties based on the number of subgoals (hard games have more than 5 subgoals),
where we observe that \ours retains a 13.3\% higher progress rate.
These results lend strong support our core claim,
that the co-evolution of exploration and world model adaptation in our approach enhances both exploration and planning in LLM agents.

\subsection{RQ1: Ablation of Hierarchical Architecture}
To evaluate the effectiveness of CoEx's hierarchical Planner-Actor architecture, we conduct an ablation study using HiAgent as our comparison baseline. 
This provides a principled ablation by comparing against a well-engineered monolithic alternative rather than artificially dismantling CoEx's integrated components. HiAgent effectively collapses our explicit Planner-Actor hierarchy into a single agent while maintaining subgoal-level planning.
While both methods employ memory mechanisms (HiAgent uses observation summarization, CoEx uses neurosymbolic belief states), this comparison isolates the impact of our architectural separation of planning and execution.
We evaluate HiAgent on the Gripper and Blocksworld tasks of PDDL with 100 steps, the same as CoEx.
The results in Table~\ref{tab:pddl_results} confirm the effectiveness of our hierarchical architecture.

\subsection{RQ2: Is \ours Computationally Efficient?}
To understand the computational costs of CoEx, we perform analysis measuring LLM API usage by tokens. 
We find that CoEx’s additional components incur only minimal overheads, with 70\% of computational costs occurring in the Actor, whose costs are shared with baseline methods like ReAct. 
Other components, i.e. the main Planner, first stage of belief state update (verification and q\&a), and belief synthesis, account for around 18\%, 10\%, 2\% of computation cost respectively. CoEx achieves this efficiency by invoking its planning and belief update components at the subgoal level rather than at every low-level action step, substantially reducing the overhead of additional components.

\subsection{RQ3: How does the World Model evolve in \ours?}
To understand how CoEx's world model co-evolves with the task subgoal progress, 
we perform a trajectory analysis of a 4-stage evolution of the world model for the task "put two soapbar in garbagecan" in the "picktwo" subset of ALFWorld\footnote{The trajectory is provided in Appendix~\ref{appendix:coex_evolution_example}.}.
In World Model Update 1, the executor’s confusion between soapbottle/soapbar objects is reflected in the belief state. 
Consequently, in World Model Update 2, the planner issues the subgoal to find soapbar in remaining unsearched locations, resulting in the successful location discovery of both soapbars at toilet 1. 
In World Model Update 3, the planner issues the subgoal to take soapbar 1, but encounters repeated failures from the executor, leading to the world model learning the knowledge 
"Agent was unable to take soapbar 1 from toilet 1, indicating a potential restriction or condition not met for that action". 
Finally, in World Model Update 4, the planner strategically adapts by targeting soapbar 2 instead, which the executor succeeds by dropping the soapbottle first, 
and in subsequent steps, this leads to eventual success on the task. 
This example demonstrates how CoEx advances the world model by integrating execution feedback, and strategically adapts plans based on discovered information about the environment.
We also note that, while LLM-based QA may introduce marginal inaccuracies during world model updates, as in existing monolithic agents, the distinction of \ours is that it can repair such errors in subsequent updates to its explicit belief state.

\section{Conclusion}

We study the problem of overcoming the limitations of existing LLM agents, namely, exploitation bias and limited adaptation.
We propose
\ours, enabling subgoal-driven exploration, allowing its internal world model to co-evolve with real-time observations. This is facilitated through a neurosymbolic belief state that integrates textual reasoning with symbolic memory, ensuring that the agent's understanding remains aligned with the true state of its environment. 
Our evaluations on benchmarks requiring planning and exploration, including ALFWorld, Jericho, and PDDL, show that CoEx outperforms strong baselines across diverse tasks.
 
\section*{Limitations}

 While the framework adaptivity has been tested in the benchmark scales,
  more abrupt or unpredictable environmental shifts may require a new strategy for faster adaptation.
  Also, some real-life tasks of sufficiently larger scales may require
 memory pruning and abstraction mechanisms. 

More sophisticated memory compression or enhanced real-time world-model synchronization are promisining future directions.

\bibliography{custom}

\appendix

\section{Appendix}

\onecolumn
\subsection{Algorithms}
\label{appendix:algos}
We present the algorithms detailing the operation of \ours.
Algorithm 1 describes the main loop that orchestrates the Planner, Actor, and belief state updates. Algorithm 2 describes the Actor's subgoal execution and the two-stage verification and synthesis process of the belief state update mechanism.

\begin{algorithm*}[t]
\caption{\ours Architecture} 
\label{alg:mainalgo} 
{\footnotesize
\begin{algorithmic}[1]
\State \textbf{Variables:}
\State \hspace{0.5cm} $\textmemk$: structured textual memory
\State \hspace{0.5cm} $\symbmemk$: symbolic memory
\State \hspace{0.5cm} $\plannerhk$: planner history
\State \hspace{0.5cm} $\subgoalk$: subgoal description (text)
\State \hspace{0.5cm} $\subepisodetrajk$: sub-episode trajectory (trace)
\State \hspace{0.5cm} $k$: main step counter
\State \textbf{Initialize:}
\State \hspace{0.5cm} $k \leftarrow 0$
\State \hspace{0.5cm} $l_0 \leftarrow []$
\State \hspace{0.5cm} $m_0 \leftarrow \text{InitializeSymbolicMemory}(\text{Env.GetInitialObs}())$
\State \hspace{0.5cm} $H_0 \leftarrow []$
\State \textbf{Input:} Environment $\env$, Planner $\planner$, actor $\actor$, VerificationAndSynthesisModule $v$

\While{task not complete and $k < \text{max\_total\_steps}$}
   \State $e_k \leftarrow \planner(H_k)$ \Comment{Planner generates next subgoal $e_k$ conditioned on belief state history}\label{lst:line:subgoal_gen}

    \State $(\varepsilon_k , m_{k+1}) \leftarrow \actor.\text{ExecuteSubgoalEpisode}(e_k)$ \Comment{Run subgoal-episode}\label{lst:line:subgoal_episode}

    \State $l_{k+1} \leftarrow v.\text{BeliefUpdate}(b_k, m_{k+1}, \varepsilon_k, e_k)$ \Comment{Verification and synthesis for belief update}

    \State $H_{k+1} \leftarrow \plannerhk \cup \{(\textmemk,\symbmemk))\}$ \Comment{Add new belief state to planner history}
    \State $k \leftarrow k + 1$
\EndWhile

\end{algorithmic}
}
\end{algorithm*}
\begin{algorithm*}[t]
\caption{\ours Subroutines}
\label{alg:hierarchical_hypoagent_direct_subgoal}
{\footnotesize
\begin{algorithmic}[1]
\Procedure{ExecuteSubgoalEpisode}{$e_k$}\label{lst:line:subgoal_execution}
    \State \textbf{Initialize:} $t \leftarrow 0$, $\text{status} \leftarrow \text{"running"}$, $H_{\text{exec}} \leftarrow \text{ExecutorInitHistory}(e_k)$
    \State \textbf{Param:} $\text{max\_sub\_steps}$
    \While{$t < \text{max\_sub\_steps}$}
        \State $a_t \leftarrow \alpha(a | e_k, H_{\text{exec}})$ 
        \If{$a_t \text{ contains } \text{SUBGOAL COMPLETED}$} \State $\text{status} \leftarrow \text{"completed"}$; \textbf{break} \EndIf
        \If{$a_t \text{ contains } \text{REQUEST\_REPLAN}$} \State $\text{status} \leftarrow \text{"not completed"}$; \textbf{break} \EndIf
        \State $o \leftarrow \text{Env.Step}(a_t)$
        \State $H_{\text{exec}} \leftarrow H_{\text{exec}} \cup \{(a_t, o)\}$
        \State $t \leftarrow t + 1$
    \EndWhile
    \State \Return $H_{\text{exec}}$
\EndProcedure

\bigskip
\Procedure{BeliefUpdate}{$b_{\text{prev}}, m_{\text{curr}}, \varepsilon, e, \text{status}$}
    \State $qa\_results \leftarrow v.\text{Verification}(b_{\text{prev}}, m_{\text{curr}}, \varepsilon, e)$
    \State $l_{\text{new}} \leftarrow v.\text{Synthesis}(b_{\text{prev}}, m_{\text{curr}}, qa\_results, e)$
    \State \Return $l_{\text{new}}$
\EndProcedure

\end{algorithmic}
}
\end{algorithm*}
\subsection{Implementation Details of Symbolic Memory}
\label{sec:symbmem_details}
We use use Gemini 2.5 pro preview to implement the symbolic memory in Python using example trajectories similar to \citet{tang2024worldcoder},
with minor finetuning of the code through manual revision.

\subsubsection{Symbolic Memory Python Implementation}
\label{appendix:symb_mem_impl_example}
We provide examples of the Python implementations of symbolic memory for the PDDL blocksworld task.

\lstdefinestyle{mypythonstyle}{
    language=Python,
    basicstyle=\ttfamily\footnotesize,
    keywordstyle=\color{blue}\bfseries,
    commentstyle=\color{green!60!black}\itshape,
    stringstyle=\color{purple},
    numberstyle=\tiny\color{gray},
    numbers=left,
    numbersep=5pt,
    tabsize=4,
    extendedchars=true,
    breaklines=true,                
    breakatwhitespace=true,         
    prebreak=\raisebox{0ex}[0ex][0ex]{\ensuremath{\hookleftarrow}}, 
    showstringspaces=false,         
    emph={self,__init__}, emphstyle=\color{red!80!black}, 
    captionpos=b,                   
    escapeinside={(*@}{@*)}         
}
\lstset{style=mypythonstyle} 

\begin{tcolorbox}[
    enhanced jigsaw,
    drop shadow=black!50!white,
    title=Python Implementation of Symbolic Memory (PDDL - Base Class),
    breakable, 
    fonttitle=\bfseries\large 
    ]

\begin{lstlisting}[caption={SimpleSymbolicMemory Base Class in Python}, label={lst:simple_symbolic_memory_base}]
import re

class SimpleSymbolicMemory:
    def __init__(self, domain_name):
        self.domain_name = domain_name
        self.predicates = set()
        self.holding = {}
        self.agent_location = None 
        self.step = 0

    def update_memory(self, observation: str, last_action: str = None):
        """
        Updates the memory based on the observation.
        This base method does nothing; subclasses must override it.
        """
        self.step += 1
        print(f"\n--- Step {self.step} ---")
        print(f"Action: {last_action}")
        print(f"Observation: {observation.strip()}")
        print(f"Predicates BEFORE update: {sorted(list(self.predicates))}")
        print(f"Holding BEFORE update: {self.holding}")
        if self.agent_location is not None:
             print(f"Location BEFORE update: {self.agent_location}")

        self._parse_and_update(observation, last_action)

        print(f"Predicates AFTER update: {sorted(list(self.predicates))}")
        print(f"Holding AFTER update: {self.holding}")
        if self.agent_location is not None:
             print(f"Location AFTER update: {self.agent_location}")

    def _parse_and_update(self, observation: str, last_action: str = None):
        # Needs implementation in subclass
        raise NotImplementedError

    def _add_predicate(self, predicate_str):
        self.predicates.add(predicate_str)

    def _remove_predicate(self, predicate_str):
        self.predicates.discard(predicate_str)

    def _remove_predicates_about(self, *args):
        to_remove = set()
        for pred in self.predicates:
            for arg in args:
                if f"({arg}," in pred or f",{arg})" in pred or f"({arg})" in pred:
                     to_remove.add(pred)
        self.predicates -= to_remove

    def _clear_holding(self, manipulator):
         if manipulator in self.holding:
             held_item = self.holding[manipulator]
             if held_item:
                 self._remove_predicates_about(
                 f"holding({manipulator},{held_item})")
             self.holding[manipulator] = None

    def _set_holding(self, manipulator, item):
        self._clear_holding(manipulator) # Ensure manipulator wasn't holding something else
        self.holding[manipulator] = item

    def get_planning_summary(self) -> str:
        lines = [f"### {self.domain_name.upper()} Memory Summary (Step {self.step}) ###"]
        if self.agent_location:
            lines.append(f"Agent Location: {self.agent_location}")
        lines.append(f"Holding: {self.holding}")
        lines.append("State:")
        if self.predicates:
            for pred in sorted(list(self.predicates)):
                lines.append(f"  - {pred}")
        else:
            lines.append("  (None)")
        lines.append("### END SUMMARY ###")
        return "\n".join(lines)

    def __str__(self):
        return self.get_planning_summary()
\end{lstlisting}
\end{tcolorbox}
\lstdefinestyle{mypythonstyle}{
    language=Python,
    basicstyle=\ttfamily\footnotesize, 
    keywordstyle=\color{blue}\bfseries,
    commentstyle=\color{green!60!black}\itshape,
    stringstyle=\color{purple},
    numberstyle=\tiny\color{gray},
    numbers=left,
    numbersep=5pt,
    tabsize=4,
    extendedchars=true,
    breaklines=true,                
    breakatwhitespace=true,         
    prebreak=\raisebox{0ex}[0ex][0ex]{\ensuremath{\hookleftarrow}}, 
    showstringspaces=false,         
    emph={self,__init__}, emphstyle=\color{red!80!black}, 
    captionpos=b,                   
    escapeinside={(*@}{@*)}         
}
\lstset{style=mypythonstyle} 

\begin{tcolorbox}[
    enhanced jigsaw,
    drop shadow=black!50!white,
    title=Python Implementation of Symbolic Memory (PDDL - BlocksWorld),
    breakable, 
    fonttitle=\bfseries\large 
    ]

\begin{lstlisting}[caption={BlocksWorldSymbolicMemory in Python}, label={lst:blocksworld_memory}]
import re

from .base import SimpleSymbolicMemory


class BlocksWorldSymbolicMemory(SimpleSymbolicMemory):
    def __init__(self):
        super().__init__("BlocksWorld")
        self.holding = {'arm': None}

    def _parse_and_update(self, observation: str, last_action: str = None):
        # 1. Determine Holding State
        held_block = None
        is_arm_empty = True
        holding_match = re.search(r"You are holding (b\d+)", observation)
        if holding_match:
            held_block = holding_match.group(1).lower()
            is_arm_empty = False
        elif re.search(r"(Robot|Your|The)\s+arm\s+is\s+empty", observation, re.IGNORECASE):
            is_arm_empty = True
        else:
            held_block = self.holding.get('arm')
            is_arm_empty = (held_block is None)

        
        self.holding['arm'] = held_block

        # 2. Gather Observed Relations
        observed_on = {m.group(1).lower(): m.group(2).lower()
                       for m in re.finditer(r"(b\d+) is on (b\d+)", observation, re.IGNORECASE)}
        observed_on_table = {m.group(1).lower()
                             for m in re.finditer(r"(b\d+) is on the table", observation, re.IGNORECASE)}
        observed_clear = {m.group(1).lower()
                          for m in re.finditer(r"(b\d+) is clear", observation, re.IGNORECASE)}
        observed_not_clear = {m.group(1).lower()
                              for m in re.finditer(r"(b\d+) is not clear", observation, re.IGNORECASE)}

        all_mentioned = (set(observed_on.keys()) | set(observed_on.values()) |
                         observed_on_table | observed_clear | observed_not_clear)
        if held_block: all_mentioned.add(held_block)

        # 3. Rebuild State for Mentioned Blocks
        new_predicates = set()

        new_predicates.add("arm_empty" if is_arm_empty else "arm_not_empty")

        blocks_underneath = set(observed_on.values())

        for block in all_mentioned:
            pos_set = False
            if block == held_block:
                pos_set = True
            elif block in observed_on:
                new_predicates.add(f"on({block},{observed_on[block]})")
                pos_set = True
            elif block in observed_on_table:
                if block not in observed_on.values():
                    new_predicates.add(f"on_table({block})")
                    pos_set = True

            clear_set = False
            if block in observed_not_clear:
                new_predicates.add(f"not_clear({block})")
                clear_set = True
            elif block == held_block:
                new_predicates.add(f"clear({block})")
                clear_set = True
            elif block in observed_clear:
                new_predicates.add(f"clear({block})")
                clear_set = True
            elif block in blocks_underneath:
                new_predicates.add(f"not_clear({block})")
                clear_set = True
            else:
                 if block != held_block:
                     new_predicates.add(f"clear({block})")
                     clear_set = True

        # 4. Update Memory
        predicates_to_remove = set()
        predicates_to_remove.add("arm_empty")
        predicates_to_remove.add("arm_not_empty")
        for block in all_mentioned:
            for pred in self.predicates:
                 if re.search(rf'\b{re.escape(block)}\b', pred):
                      if pred.startswith(('on(', 'on_table(', 'clear(', 'not_clear(')):
                           predicates_to_remove.add(pred)
        self.predicates -= predicates_to_remove

        self.predicates.update(new_predicates)
\end{lstlisting}
\end{tcolorbox}

\subsection{Prompts}
\label{appendix:prompts}

\subsubsection{\ours Planner Prompts}
\label{appendix:planner_prompts}
\begin{tcolorbox}[enhanced jigsaw,drop shadow=black!50!white,title=System Prompt (ALFWorld)]
You are Alfred, an agent for the ALFWorld household environment. You will be given a task to complete, and you complete the task by breaking down the task into a task consisting of subgoals. You can use the \texttt{EXECUTE\_SUBGOAL} action to execute a subgoal, which will be delegated to a subgoal executor agent, and you will receive feedback on the subgoal execution. When executing subgoals, please be detailed, and include the necessary relevant information about the status of the task along with the subgoal.
\vspace{\medskipamount}

\textbf{Generating Plan}\\
Whenever you generate an entirely new plan, or you change the plan with different subgoals or new order of subgoals, you must include \texttt{FULL PLAN} in the response, followed by all of the steps in the new plan. Example:
\begin{quote}
\texttt{...}\\
\texttt{FULL PLAN}\\
\texttt{Subgoals:}\\
\texttt{1. Go to ...}\\
\texttt{2. Take ...}\\
\texttt{3. Use ...}\\
\texttt{4. Go to ...}\\
\texttt{5. ...}
\end{quote}
%\vspace{\medskipamount}

\textbf{Processing Subgoal Feedback:}
\begin{itemize}[label=-, leftmargin=*,itemsep=0pt]
    \item After a subgoal attempt, you will receive an \texttt{'analysis\_feedback'} message (role: assistant).
    \item \textbf{Parse the \texttt{JSON} content} of this message.
    \item Use the \texttt{new\_belief} field to understand the current world state before planning the next subgoal.
\end{itemize}
%\vspace{\medskipamount}

\textbf{\texttt{EXECUTE\_SUBGOAL} Action Format:}\\
You MUST issue subgoals using the following multi-line format precisely:
\begin{quote}
\texttt{EXECUTE\_SUBGOAL[}\\
\texttt{\ \ DESC: <Clear, natural language description of the specific subgoal>}\\
\texttt{\ \ SEARCH\_LOCATIONS: [<loc1>, <loc2>, ...] \# Include ONLY when we need to search for an object, otherwise omit or null}
\end{quote}
\vspace{\medskipamount}

\textbf{Generating \texttt{SEARCH\_LOCATIONS} (Crucial for Find/Take Subgoals):}\\
When the \texttt{DESC} is "\texttt{Find and take [object]}", you MUST:
\begin{enumerate}[label=\arabic*), leftmargin=*,itemsep=0pt]
    \item Read the \texttt{new\_belief} text from the most recent \texttt{analysis\_feedback} message.
    \item Identify \textbf{all receptacle IDs} mentioned as existing in the room within that belief text (e.g., \texttt{cabinet 1}, \texttt{fridge 1}, \texttt{countertop 1}).
    \item Create a list of these known receptacle IDs. Prioritize likely locations if possible.
    \item Populate the \texttt{SEARCH\_LOCATIONS:} field with this \textbf{exact list}. Example: \texttt{SEARCH\_LOCATIONS: [cabinet 1, fridge 1, countertop 1]}
\end{enumerate}
\vspace{\medskipamount}

Below are examples of a similar task:\\
\texttt{\{\{task exemplars\}\}}
\end{tcolorbox}

\vspace{\bigskipamount}

\begin{tcolorbox}[enhanced jigsaw,drop shadow=black!50!white,title=Instance Prompt (ALFWorld)]
\texttt{\{\{task\_room\}\}\{\{task\}\}}
\vspace{\smallskipamount}

Exploration strategies:
\begin{itemize}[label=-, leftmargin=*]
    \item When searching for an object, try to search in the most likely location first, and if not found, try to expand the search to more unlikely locations.
\end{itemize}
\vspace{\medskipamount}

\end{tcolorbox}
\begin{tcolorbox}[enhanced jigsaw,drop shadow=black!50!white,title=System Prompt (Jericho)]
You are an agent playing a text-based adventure game. You will be given a task in the game to complete, and you complete the task by breaking down the task into a task consisting of subgoals. You can use the \texttt{EXECUTE\_SUBGOAL} action to execute a subgoal, which will be delegated to a subgoal executor agent, and you will receive feedback on the subgoal execution. When executing subgoals, please be detailed, and include the necessary relevant information about the status of the task along with the subgoal.
\vspace{\medskipamount}

\textbf{Generating Plan}\\
Whenever you generate an entirely new plan, or you change the plan with different subgoals or new order of subgoals, you must include \texttt{FULL PLAN} in the response, followed by all of the steps in the new plan. When doing so you MUST include the \texttt{FULL PLAN} in the response, followed by all of the steps in the new plan.
%\vspace{\smallskipamount}
Example:
\begin{quote}
\texttt{...}\\
\texttt{FULL PLAN}\\
\texttt{Subgoals:}\\
\texttt{1. Go to ...}\\
\texttt{2. Take ...}\\
\texttt{3. Use ...}\\
\texttt{4. Go to ...}\\
\texttt{5. ...}
\end{quote}
%\vspace{\medskipamount}

\textbf{Processing Subgoal Feedback:}
\begin{itemize}[label=-, leftmargin=*,itemsep=0pt]
    \item After a subgoal attempt, you will receive an \texttt{'analysis\_feedback'} message (role: assistant).
    \item \textbf{Parse the \texttt{JSON} content} of this message.
    \item Use the \texttt{new\_belief} field to understand the current world state before planning the next subgoal.
\end{itemize}
\vspace{\medskipamount}

\textbf{\texttt{EXECUTE\_SUBGOAL} Action Format:}\\
You MUST issue subgoals using the following multi-line format precisely:
\begin{quote}
\texttt{EXECUTE\_SUBGOAL[}\\
\texttt{\ \ DESC: <Clear, natural language description of the specific subgoal>}\\
\texttt{]}
\end{quote}
\vspace{\medskipamount}

\textbf{Checking valid actions}
\begin{itemize}[label=-, leftmargin=*,itemsep=0pt]
    \item If you are unsure about which actions can be taken, make sure to use the \texttt{'check valid actions'} command.
\end{itemize}
\vspace{\medskipamount}

Below are examples of a similar task:\\
\texttt{\{\{task exemplars\}\}}
\end{tcolorbox}

\vspace{\bigskipamount}

\begin{tcolorbox}[enhanced jigsaw,drop shadow=black!50!white,title=Instance Prompt (Jericho)]
\texttt{Goal: \{\{ goal \}\}}\\
\vspace{\smallskipamount}
\texttt{\{\{ initial\_observation \}\}}
\end{tcolorbox}
\begin{tcolorbox}[
    enhanced jigsaw,
    drop shadow=black!50!white,
    title=System Prompt (PDDL),
    breakable
    ]
You are an agent carrying out a planning task. You will be given a task to complete, and you complete the task by breaking down the task into subgoals. You can use the \texttt{EXECUTE\_SUBGOAL} action to execute a subgoal, which will be delegated to a subgoal executor agent, and you will receive feedback on the subgoal execution. When executing subgoals, please be detailed, and include the necessary relevant information about the status of the task along with the subgoal.
\vspace{\medskipamount}

\textbf{Generating Plan}\\
Whenever you generate an entirely new plan, or you change the plan with different subgoals or new order of subgoals, you must include \texttt{FULL PLAN} in the response, followed by all of the steps in the new plan. When doing so you MUST include the \texttt{FULL PLAN} in the response, followed by all of the steps in the new plan.
%\vspace{\smallskipamount}
Example:
\begin{quote}
\texttt{...}\\
\texttt{FULL PLAN}\\
\texttt{Subgoals:}\\
\texttt{1. Go to ...}\\
\texttt{2. Take ...}\\
\texttt{3. Use ...}\\
\texttt{4. Go to ...}\\
\texttt{5. ...}
\end{quote}
%\vspace{\medskipamount}

\textbf{Processing Subgoal Feedback:}
\begin{itemize}[label=-, leftmargin=*,itemsep=0pt]
    \item After a subgoal attempt, you will receive an \texttt{'analysis\_feedback'} message (role: assistant).
    \item \textbf{Parse the \texttt{JSON} content} of this message.
    \item Use the \texttt{new\_belief} field to understand the current world state before planning the next subgoal.
\end{itemize}
\vspace{\medskipamount}

\textbf{\texttt{EXECUTE\_SUBGOAL} Action Format:}\\
You MUST issue subgoals using the following multi-line format precisely:
\begin{quote}
\texttt{EXECUTE\_SUBGOAL[}\\
\texttt{\ \ DESC: <Clear, natural language description of the specific subgoal>}\\
\texttt{]}
\end{quote}
\vspace{\medskipamount}

Below are examples of a similar task:\\
\texttt{\{\{task exemplars\}\}}
\end{tcolorbox}

\vspace{\bigskipamount}

\begin{tcolorbox}[enhanced jigsaw,drop shadow=black!50!white,title=Instance Prompt (PDDL)]
Goal: \\
\texttt{\{\{ goal \}\}}
\vspace{\medskipamount}
\\Initial Observation: \\
\texttt{\{\{ initial\_observation \}\}}
\vspace{\medskipamount}
\\Begin by carefully summarizing the target goal state, and then generate the plan to achieve the target goal.
\end{tcolorbox}

\clearpage
\subsubsection{\ours Subgoal Prompts}
\label{appendix:subgoal_prompts}
The example below shows the actor prompt for PDDL. 
For PDDL, domain-specific instructions are provided for each domain,
while for ALFWorld and Jericho, the instructions are shared for all tasks.
\newcommand{\promptseparator}{
  \par\noindent\makebox[\linewidth]{\color{gray}\leaders\hrule height 0.4pt\hfill}\par
}
\begin{tcolorbox}[
    enhanced jigsaw,
    drop shadow=black!50!white,
    title=System Prompt,
    breakable
    ]
You are a master in planning. You will be given a subgoal to complete.
\vspace{\medskipamount}
Think step-by-step. Output your thought process followed by the command in markdown 
\vspace{\medskipamount}

When you execute a command, you will receive a response from the game.
\begin{itemize}[label=\textbullet, itemsep=0pt, leftmargin=*]
    \item If the action is successful, you will receive the updated state of the world.
    \item If the action is unsuccessful, you will receive a message indicating the failure: "\texttt{The action is not valid and therefore takes no effect. Please check valid actions.}"
\end{itemize}
\vspace{\medskipamount}

\begin{itemize}[label=\textbullet, itemsep=0pt, leftmargin=*]
    \item If you believe the subgoal is achieved based on the game's response, output the action \texttt{'SUBGOAL COMPLETED'} instead of a game command.
    \item If you get stuck, cannot proceed, or believe the subgoal is impossible, output \texttt{'REQUEST\_REPLAN[<Reason for failure>]'}.
\end{itemize}
\end{tcolorbox}

\vspace{\bigskipamount}

\begin{tcolorbox}[enhanced jigsaw,drop shadow=black!50!white,title=Instance Prompt]
\texttt{\{\{domain\_instructions\}\}}
\vspace{\medskipamount}

Think step-by-step about your plan and the expected outcome before issuing a command.
Format your response with your thought process, followed by the command in markdown backticks.
Do not issue multiple commands at once. If you issue the first command, wait for the result before issuing the next command.
\vspace{\medskipamount}

Example Format:\\
\texttt{\{\{example\_format\}\}}
\vspace{\bigskipamount}

\promptseparator
Your Assigned Subgoal: \texttt{\{\{ subgoal \}\}}
\promptseparator
Your Current State:\\
\texttt{\{\{ location \}\}}
\promptseparator
\vspace{\medskipamount}
Execute the next command towards the subgoal, or output \texttt{SUBGOAL COMPLETED} or \texttt{REQUEST\_REPLAN[...]}
\end{tcolorbox}

\clearpage
\subsubsection{\ours Verification and Synthesis Prompts}
\label{appendix:vas_prompts}
The same verification and synthesis prompts are used for all tasks.
The \texttt{context} variable is constructed by concatenating: 
the subgoal text $\subgoalk$,
the raw action and observation trace from the subgoal trajectory $\subepisodetrajk$, 
and the most recent symbolic memory $\symbmemk$.
The LLM is prompted with the system prompt as well as an instance prompt per each question, generating the answers to the questions.
\begin{tcolorbox}[
    enhanced jigsaw,
    drop shadow=black!50!white,
    title=System Prompt for Verification Stage,
    breakable
    ]
You are an analytical assistant answering specific questions about an agent's execution trace. Provide a clear answer (Yes/No/Uncertain/Specific Value) and a brief justification based \textit{only} on the provided context.
\vspace{\medskipamount}

The assistant will be asked one of the following types of questions:
\begin{itemize}[label=\textbullet, leftmargin=*, itemsep=0pt]
    \item "Did the subgoal '\texttt{<<subgoal>>}' contribute positively towards the main goal based on the trace?"
    \item "Did the agent successfully navigate to the intended location or interact with the intended object?"
    \item "Were there any errors (e.g., 'You can't do that', 'I don't understand') or loops?"
    \item "Did the agent's inventory change as expected?"
    \item "Based ONLY on the execution trace, what are the 1-3 most important new facts learned, errors encountered, or surprising outcomes observed during this subgoal attempt? List them concisely or state 'None'."
\end{itemize}
\end{tcolorbox}

\vspace{\bigskipamount}

\begin{tcolorbox}[enhanced jigsaw,drop shadow=black!50!white,title=Instance Prompt for Verification Stage]
\texttt{Based ONLY on the provided context below, answer the following question.}\\
\texttt{ }\\
\texttt{CONTEXT:}\\
\texttt{\{\{context\}\}}\\
\texttt{ }\\
\texttt{QUESTION: \{\{question\}\}}\\
\texttt{ }\\
\texttt{ANSWER (e.g., Yes/No/Uncertain/Value): [Your Answer]}\\
\texttt{JUSTIFICATION: [Your Brief Reasoning]}
\vspace{\medskipamount}

\end{tcolorbox}

The synthesis stage takes the outputs of the Verification QA stage, 
along with the previous belief state $b_{k-1}$,
symbolic memory $\symbmemk$, the subgoal text $\subgoalk$,
to generate the new belief state update.

\begin{tcolorbox}[
    enhanced jigsaw,
    drop shadow=black!50!white,
    title=System Prompt for Synthesis Stage,
    breakable
    ]
You are a high-level planner agent. Based on the previous belief, current memory state, latest plan, and analysis of the last subgoal execution (Q\&A), decide the next course of action.
\vspace{\medskipamount}

Generate ONLY the following:
\begin{enumerate}[label=\arabic*., leftmargin=*, itemsep=0pt]
    \item A concise status line reflecting the current progress relative to the plan (starting with "\texttt{Status: }").
    \item and the justification for the status line.
    \item A list of concise new facts learned or hypotheses formed about the environment/task based \textit{only} on the \textit{last} subgoal's execution, especially failures or unexpected outcomes. Focus on actionable insights or constraints.
\end{enumerate}
\vspace{\medskipamount}

Respond ONLY with a valid JSON object containing the keys "\texttt{status\_line}" (string), "\texttt{justification}" (string), and "\texttt{learned\_facts}" (list of strings, can be empty).
\end{tcolorbox}

\vspace{\bigskipamount}

\begin{tcolorbox}[enhanced jigsaw,drop shadow=black!50!white,title=Instance Prompt for Synthesis Stage]
Previous Belief:\\
\texttt{\{\{ previous\_belief \}\}}
\vspace{\medskipamount}

Current Symbolic Memory Summary:\\
\texttt{\{\{ memory\_summary \}\}}
\vspace{\medskipamount}

Latest Overall Plan:\\
\texttt{\{\{ latest\_plan \}\}}
\vspace{\medskipamount}

Last Subgoal Attempted: \texttt{\{\{ subgoal \}\}}\\
\vspace{\medskipamount}

Subgoal Execution Q\&A Analysis:\\
\texttt{\{\{ qa\_summary \}\}}
\vspace{\bigskipamount}

\textbf{Instructions:}
\begin{enumerate}[label=\arabic*., leftmargin=*, itemsep=0pt]
    \item \textbf{Generate \texttt{status\_line}:} Create a single sentence starting with "\texttt{Status: }" that reflects the '\texttt{Last Subgoal Outcome}' and progress relative to the '\texttt{Latest Overall Plan}'.
    \item \textbf{Provide \texttt{justification}:} Briefly explain the reasoning for the status line.
    \item \textbf{Generate \texttt{learned\_facts}:} Based \textit{only} on the Q\&A and outcome of the \textit{last} subgoal, list any new, concise, and actionable facts learned or hypotheses formed (e.g., "\texttt{Confirmed: Kettle is in Cabinet.}", "\texttt{Error: Agent tried incorrect action 'open fridge drawer.'}"). Output an empty list \texttt{[]} if nothing significant was learned.
\end{enumerate}
\vspace{\medskipamount}

Respond ONLY with a valid JSON object containing "\texttt{status\_line}", "\texttt{justification}", and "\texttt{learned\_facts}".
\end{tcolorbox}

\subsection{Example of World Model Evolution}
Below we provide the trajectory showing a 4-stage evolution of the world model in \ours, 
for the ALFWorld task "put two soapbar in garbagecan". 
This example demonstrates how \ours integrates execution feedback into its belief state, and how the Planner strategically adapts its subgoals based on this newly discovered information about the environment.
\label{appendix:coex_evolution_example}
\begin{tcolorbox}[
    enhanced jigsaw,
    drop shadow=black!50!white,
    title=World Model Update 1,
    breakable
    ]
\textbf{Subgoal Generation and Execution:}
\vspace{\medskipamount}

\textbf{Planner Reasoning:} ``Starting with the first subgoal. Likely locations for soapbar: countertop, cabinets, towelholders, sinkbasins''

\textbf{Subgoal:} Find and take the first soapbar (Search loc: [countertop 1, cabinet 1-4, handtowelholder 1-2, sinkbasin 1-2])

\textbf{Subgoal Result:} \textcolor{red}{\textbf{\xmark}} Took wrong object

\vspace{\medskipamount}
\textbf{Updated Belief State (Abbreviated):}
\begin{itemize}[leftmargin=*, itemsep=0pt]
    \item \textbf{Symbolic Memory:} \texttt{Agent: ``holding soapbottle 1''}
    \item \textbf{Semantic:} ``Error: Agent took soapbottle 1 instead of the target soapbar''
\end{itemize}
\end{tcolorbox}

\vspace{\bigskipamount}

\begin{tcolorbox}[
    enhanced jigsaw,
    drop shadow=black!50!white,
    title=World Model Update 2,
    breakable
    ]
\textbf{Subgoal Generation and Execution:}
\vspace{\medskipamount}

\textbf{Planner Reasoning:} ``Search the remaining unexplored locations: garbagecan, sinkbasins, toilet, toiletpaperhanger, and towelholder''

\textbf{Subgoal:} Find and take the first soapbar (Search loc: [garbagecan 1, sinkbasin 1, sinkbasin 2, toilet 1, toiletpaperhanger 1, towelholder 1])

\textbf{Subgoal Result:} \textcolor{green}{\textbf{\cmark}} Found soapbars

\vspace{\medskipamount}
\textbf{Updated Belief State (Abbreviated):}
\begin{itemize}[leftmargin=*, itemsep=0pt]
    \item \textbf{Symbolic Memory:} \texttt{``toilet 1: contains=[soapbar 1, soapbar 2]''}
    \item \textbf{Semantic:} ``Confirmed: Soapbar 1 and soapbar 2 are located at toilet 1''
\end{itemize}
\end{tcolorbox}

\vspace{\bigskipamount}

\begin{tcolorbox}[
    enhanced jigsaw,
    drop shadow=black!50!white,
    title=World Model Update 3,
    breakable
    ]
\textbf{Subgoal Generation and Execution:}
\vspace{\medskipamount}

\textbf{Planner Reasoning:} ``Since I have confirmed that soapbar 1 is currently in toilet 1, I will first take the soapbar from toilet 1''

\textbf{Subgoal:} Take soapbar 1 from toilet 1 (Search loc: null)

\textbf{Subgoal Result:} \textcolor{red}{\textbf{\xmark}} Repeatedly failed

\vspace{\medskipamount}
\textbf{Updated Belief State (Abbreviated):}
\begin{itemize}[leftmargin=*, itemsep=0pt]
    \item \textbf{Symbolic Memory:} \texttt{Agent: ``holding soapbottle 1''}
    \item \textbf{Semantic:} ``Error: Agent was unable to take soapbar 1 from toilet 1, indicating a potential restriction or condition not met for that action''
\end{itemize}
\end{tcolorbox}

\vspace{\bigskipamount}

\begin{tcolorbox}[
    enhanced jigsaw,
    drop shadow=black!50!white,
    title=World Model Update 4,
    breakable
    ]
\textbf{Subgoal Generation and Execution:}
\vspace{\medskipamount}

\textbf{Planner Reasoning:} ``Since the soapbars are confirmed to be located at toilet 1, I will attempt to take soapbar 2 from toilet 1 directly, as the previous attempts to take soapbar 1 were unsuccessful''

\textbf{Subgoal:} Take soapbar 2 from toilet 1 (Search loc: null)

\textbf{Subgoal Result:} \textcolor{green}{\textbf{\cmark}} Success

\vspace{\medskipamount}
\textbf{Updated Belief State:}
\begin{itemize}[leftmargin=*, itemsep=0pt]
    \item \textbf{Symbolic Memory:} \texttt{Agent: ``holding soapbar 2'', toilet 1: ``contains=[soapbar 1, soapbottle 1]''}
    \item \textbf{Semantic:} ``Subgoal completed; soapbar 2 successfully taken from toilet 1''
\end{itemize}
\end{tcolorbox}

\subsection{Compatibility of Symbolic Memory with Symbolic Solvers}
Our symbolic memory maintains accurate object-oriented state tracking, and in PDDL the representation maps to formal predicates compatible with a symbolic solver. In ALFWorld and Jericho, a symbolic solver would be incompatible due to environment complexity. Our method handles such complexity by integrating an LLM-based hierarchical planner with symbolic memory and semantic belief state.

\end{document}